\documentclass[12pt,a4paper,twoside]{article}
\usepackage[margin=1in,headheight=15pt]{geometry}

\usepackage{amsmath,amssymb,amsthm,amsfonts,amsthm}
\usepackage{placeins}

\newcommand{\smallbullet}{\vcenter{\hbox{\tiny$\bullet$}}}
\usepackage{xcolor,colortbl,caption}
\usepackage[FIGTOPCAP]{subfigure}

\usepackage{natbib}
\newcommand\citepos[1]{\citeauthor{#1}'s (\citeyear{#1})}
\usepackage[breaklinks=true]{hyperref}
\usepackage{breakcites}

\usepackage{fancyhdr}
\newcommand{\Enc}{\mbox{\tt Enc}}
\newcommand{\Dec}{\mbox{\tt Dec}}

\usepackage{fancyvrb}
\DefineVerbatimEnvironment{Highlighting}{Verbatim}{commandchars=\\\{\}}
\usepackage{framed}
\definecolor{shadecolor}{RGB}{248,248,248}
\newenvironment{Shaded}{\begin{snugshade}}{\end{snugshade}}
\newcommand{\KeywordTok}[1]{\textcolor[rgb]{0.13,0.29,0.53}{\textbf{{#1}}}}

\newcommand{\DecValTok}[1]{\textcolor[rgb]{0.00,0.00,0.81}{{#1}}}

\newcommand{\StringTok}[1]{\textcolor[rgb]{0.31,0.60,0.02}{{#1}}}

\newcommand{\NormalTok}[1]{{#1}}

\title{A review of homomorphic encryption and software tools for encrypted statistical machine learning}
\author{Louis~J.~M.~Aslett, Pedro~M.~Esperan\c{c}a and Chris~C.~Holmes  \\ {\small Department of Statistics, University of Oxford}}
\date{}

\begin{document}
\maketitle

\begin{abstract}
Recent advances in cryptography promise to enable secure statistical computation on encrypted data, whereby a limited set of operations can be carried out without the need to first decrypt.  We review these \emph{homomorphic encryption schemes} in a manner accessible to statisticians and machine learners, focusing on pertinent limitations inherent in the current state of the art.  These limitations restrict the kind of statistics and machine learning algorithms which can be implemented and we review those which have been successfully applied in the literature.  Finally, we document a high performance R package implementing a recent homomorphic scheme in a general framework.

\vspace{5mm}
\end{abstract}

\section{Introduction}

The extensive use of private and personally identifiable information in modern statistical (and machine learning) applications can present an obstacle to individuals contributing their data to research.  As just one example, when considering contribution to biobanks \citet{Kaufman09} reported 90\% of respondents had privacy concerns.  Addressing these concerns is paramount if the participation rate in biomedical and genetic research is to be increased, especially for government and industry where public trust is lower \citep{Kaufman09}.  Indeed, industry is on the brink on embarking on biomedical applications on a scale never before witnessed via the impending wave of so-called `wearable devices' such as smart watches, which present serious privacy concerns.  Companies hope to market the ability to monitor and track vital health signs round the clock, perhaps fitting classification models to alert different health concerns of interest.  However, such constrained devices will almost certainly leverage `cloud' services, uploading reams of private health diagnostics to corporate servers.  Herein, it is demonstrated how recent advances in cryptography allow individual privacy to be preserved, whilst still enabling researchers and industry to incorporate such data into statistical analyses.

Moreover, the current explosion in cloud computing platforms  promise to enable researchers and businesses to divest themselves of complex in-house compute server setups, but require one to vest all trust in the cloud provider maintaining confidentiality of the data.

One way to ensure trust in the scenarios above is through storage and disclosure of only secure, encrypted data.  Encryption is a technique whereby data, termed a \emph{message} in cryptography, is mathematically transformed using an encryption key to produce a \emph{cipher text}.  The cipher text can only easily be decrypted to reveal the original data if the corresponding decryption key is known.  Therefore, a cipher text can be stored openly without compromising privacy so long as the decryption key is kept secret.

From a data science perspective, the problem with employing cryptographic methods to improve trust is that the data must at some point be decrypted for use in a statistical analysis.  However, recent cryptography research in the areas of \emph{homomorphic} and \emph{functional} encryption are showing exciting potential to bypass this.  An encryption scheme is said to be homomorphic if certain mathematical operations can be applied directly to the cipher text in such a way that decrypting the result renders the same answer as applying the function to the original unencrypted data.

The remarkable properties of homomorphic encryption schemes are not without limitations, which typically include slow evaluation and the fact that the set of functions which can be computed in cipher text space is very restricted.  However, by understanding the constraints and restrictions it is hoped that statistics researchers can assist in the  research effort, adapting statistical techniques to be amenable to homomorphic computation by making and quantifying reasonable approximations in those situations where a traditional approach cannot be implemented homomorphically.

There are reviews and introductions to homomorphic encryption aimed at different audiences and each with a different emphasis \citep{Gentry10, vaikuntanathan2011computing, sen, silverberg2013fully}.  The aim of this paper is to provide statisticians and machine learners with sufficient background to become involved in developing methodology specifically crafted to homomorphic computation. As part of this effort we describe an accompanying high performance R package providing an easy to use reference implementation as a core contribution of this work.  In a sister publication \citep*{Part2} we present some novel statistical machine learning techniques developed to be amenable to fitting and prediction encrypted.

In Section 2 homomorphic encryption is introduced covering the salient features for statistical work without drifting too far into cryptography theory unnecessarily, although full references and resources are provided for further reading.  Section 3 reviews the statistical techniques which have been successfully implemented in the cryptography literature and existing software implementations of homomorphic schemes.  Section 4 describes a high-level easy to use software implementation available as an R package \citep{HEpkg}.

\section{Homomorphic encryption}

This section presents an introduction to homomorphic encryption with an emphasis  on details and limitations which are pertinent to applying statistics and machine learning methodology.

\subsection{Background on encryption}

An unencrypted number, $m \in M$, is referred to as a \emph{message}, while the encrypted version, $c \in C$, is the \emph{cipher text}, where $M$ and $C$ are the message space and cipher text space respectively.  Typically $M \subset \mathbb{Z}$, the integers or similar, whilst $C$ will depend on the encryption algorithm being used.  
A given encryption scheme then utilises \emph{keys} in order to map the message into a cipher text and to recover the message from a cipher text.  There are two approaches: either there is a single secret key, or there are a public and secret key.  In the single secret key scheme the same key is used to map messages to cipher texts and vice versa, so this key must be kept private at all times.  Conversely, a scheme which also has a public key uses that key to map messages to cipher texts, but uses the secret key to map back: consequently the public key can be openly disclosed.  Hereinafter, only public key schemes are considered.

Fundamentally encryption can be treated as simply a mapping which takes $m$ and a \emph{public key}, $k_p$, and produces the cipher text, $c \leftarrow \Enc(k_p, m)$.  Notationally, $\leftarrow$ is used to signify assignment rather than equality, since encryption is not necessarily a function in the mathematical sense: any fixed inputs $k_p$ and $m$ will produce many different cipher texts.  Indeed, this is a desirable property for public key encryption schemes, referred to as \emph{semantic security}: a scheme is semantically secure if knowledge of $c$ for some $m$ has vanishingly small probability of revealing further information about any other encrypted message.  Informally, this means repeated encryption of the same message $m$ will render different and seemingly unrelated cipher texts each time with high probability.  Clearly, if encryption was an injective function for fixed $k_p$, $\Enc : M \to C$, then any public key encryption scheme with a modestly sized message space could be trivially compromised.  Semantic security is achieved by introducing randomness into the cipher text which is sufficiently small not to interfere with correct decryption when in possession of $k_s$, but, as will become apparent in the sequel, this essential feature imposes a handicap on all currently known homomorphic schemes.

Conversely, decryption is a function which renders the original message, $m = \Dec(k_s, c)$.  The crucial relation satisfied by any encryption scheme is therefore: \[ m = \Dec(k_s, \Enc(k_p, m)) \quad \forall \ m \in M \]

Consequently, the security of an encryption scheme is based on the hardness of recovering $m$ given knowledge of only $c$ and $k_p$.  Some schemes are based on empirical hardness assumptions about particular problems, whilst others may rely on settings where the hardness can be rigorously proven.

This is a simplification of general cryptographic schemes, since some of the most important algorithms, such as the current industry standard Advanced Encryption Standard (AES) \citep{daemen2002design}, do not normally operate value-by-value but rather on blocks of binary data.  However, it encompasses the class of algorithms to be discussed in what follows.

\subsection{Homomorphic encryption}
\label{sec:HE}

The term \emph{homomorphic encryption} describes a class of encryption algorithms which satisfy the homomorphic property: that is certain operations, such as addition, can be carried out on cipher texts directly so that upon decryption the same answer is obtained as operating on the original messages.  In simple terms, were one to encrypt the numbers 2 and 3 separately and `add' the cipher texts, then decryption of the result would yield 5.  This is a special property not enjoyed by standard encryption schemes where decrypting the sum of two cipher texts would generally render nonsense.

More precisely, an encryption scheme is said to be homomorphic for some operations $\circ \in \mathcal{F}_M$ acting in message space (such as addition) if there are corresponding operations $\diamond \in \mathcal{F}_C$ acting in cipher text space satisfying the property:
\[ \Dec(k_s, \Enc(k_p, m_1) \diamond \Enc(k_p, m_2)) = m_1 \circ m_2 \]

For example, the simple scheme in \citet{Gentry10} describes a method where $\mathcal{F}_M = \{+, \times\}$ and $\mathcal{F}_C = \{+, \times\}$, though there is no restriction that the operations must correspond in all schemes.  For example, Paillier encryption \citep{paillier1999public} is homomorphic only for addition, with $\mathcal{F}_M = \{+\}$ but where $\mathcal{F}_C = \{\times\}$.

Note this is not a group homomorphism in the mathematical sense, since the property does not commute when starting instead from cipher texts, due to semantic security.  That is, because the same message encrypts to different cipher texts with high probability, in general: \[\Enc(k_p, m_1) \diamond \Enc(k_p, m_2) \ne \Enc(k_p, m_1 \circ m_2)\]
Moreover, generally $m_1 > m_2 \nRightarrow \Enc(k_p, m_1) > \Enc(k_p, m_2)$.  Another consequence of semantic security is that operations performed on the cipher text may increase the noise level, so that only a limited number of operations can be consecutively performed before the noise must be reduced.

The possibility of homomorphic encryption was proposed by \citet{Rivest78} and many schemes that supported either multiplication (such as RSA \citep{rivest1978method}, ElGamal \citep{elgamal1985public}, etc) or addition (such as Goldwasser-Micali \citep{goldwasser1982probabilistic}, Paillier \citep{paillier1999public}, etc) were found.  However, in many of these the number of times one could add or multiply was limited and a scheme supporting both operations simultaneously was elusive (\citet{boneh2005evaluating} came closest, allowing unlimited additions and a single multiplication).  It was not until 2009 that the three decade old problem was solved in seminal work by \citet{Gentry09}, where he showed addition, multiplication and control of the noise growth were all possible.  This sparked a cascade of work on \emph{fully} homomorphic schemes: that is, those where a theoretically unlimited number of addition and multiplication operations are possible.  This modern era of homomorphic encryption is briefly summarised in Appendix \ref{sec:modernHE}.

The advent of a scheme capable of evaluating both addition and multiplication a (theoretically) arbitrary number of times led to a surge of optimism, since then any polynomial can be computed and so the output of any suitably smooth function could in principal be arbitrarily closely approximated.  Moreover, if $M=\{0,1\}$ then addition corresponds to logical XOR, and multiplication corresponds to logical AND, which is sufficient to construct arbitrary binary circuits so that, in principle, anything which can be evaluated by a computer can be represented by an algorithm which will run on homomorphically encrypted data.  However, caution is needed here regarding practicality: performing just a 32-bit integer addition using a simple ripple-carry adder design involves 32 full adders, each requiring 3 XORs, 2 ANDs and an OR ($\equiv$ 2 XOR \& 1 AND) --- 256 fundamental operations just to add two integers, an avenue it will become clear is impractical with current homomorphic schemes.

A slightly whimsical but highly lucid and more detailed introduction to homomorphic encryption can be found in \citet{Gentry10}.  A longer introduction and background is in \citet{sen}.

\subsection{The scheme of \citet{Fan12}}
\label{sec:FandV}

To make these ideas more concrete the particular scheme of \citet{Fan12} (hereinafter \texttt{FandV}) will now be described.  A high performance, easy to use implementation of the same is a contribution of this technical report as discussed in Section \ref{sec:Rpkg}.

\texttt{FandV} is a fully homomorphic scheme where the message space accommodates representation of large subsets of $\mathbb{Z}$ (not just binary messages), and a cipher text is a pair of large polynomials.  Its security is based on the hardness of the ring Learning With Errors (LWE) problem \citep{ringlwe} which is connected to classical cryptography hardness results (such theory would be a diversion: for a short description see Appendix \ref{sec:ringlwe}).

To simplify the presentation for a statistics audience, some minor simplifying restrictions are made to the original scheme as will be explained.  The reader may safely skip to Section \ref{sec:limitations} if the following mathematical details of this example encryption scheme are not of interest.

\subsubsection{Notation}

$\mathbb{Z}_q$ is the set of integers $\{ n : n \in \mathbb{Z}, -q/2 < n \le q/2 \}$ and $[a]_q$ denotes the unique integer in $\mathbb{Z}_q$ which is equal to $a \mod q$.  $\mathbb{Z}[x]$ and $\mathbb{Z}_q[x]$ denote polynomials whose coefficients belong to $\mathbb{Z}$ and $\mathbb{Z}_q$ respectively.  Then, for a fixed value $d$, the primary objects of interest in the scheme are the polynomial rings $R=\mathbb{Z}[x] / \Phi_{2^d}(x)$ and $R_q=\mathbb{Z}_q[x] / \Phi_{2^d}(x)$, where $\Phi_{2^d}(x)=x^{2^{d-1}}+1$ is the $2^d$-th cyclotomic polynomial\footnote{In simple terms, $\Phi_d(x)$, the $d$-th cyclotomic polynomial is the polynomial which: divides $x^d-1$; does not divide $x^n-1$ for any $n<d$; has integer coefficients; and cannot be factorised.
	
For example, $\Phi_3(x) = x^2+x+1$ because $(x^2+x+1)(x-1)=x^3-1$, but it does not divide $x^2-1$ or $x-1$, it has integer coefficients and it cannot be factorised.}.  The restriction to \mbox{$2^d$-th} cyclotomic polynomials here is for the convenience of their form, the computational efficiencies of reducing a polynomial modulo this form, and for the simplicity of generating random polynomials modulo this form which satisfy ring LWE hardness results (although theoretically \texttt{FandV} can be modulo any monic irreducible polynomial).

To distinguish polynomials, they will be underscored $\underline{a} \in R_q$ if not written in functional form, $a(x)$.  Polynomial multiplication will be emphasised, $\underline{a}\cdot\underline{b}$ and all such multiplication takes place within the ring $R$.  $[\underline{a}]_q$ indicates the centred reduction above applied to each coefficient of $\underline{a}$ individually, so that $\underline{a} \in R \implies [\underline{a}]_q \in R_q$.

The randomness to be introduced for semantic security comes via the bounded discrete Gaussian distribution, defined to be the probability mass function proportional to $\exp(-x^2/(2\sigma^2))$ over the integers from $-B$ to $B$, where typically $B \approx 10\sigma$.  For the special choice of polynomial modulo $\Phi_{2^d}(x)$ above, the corresponding multivariate distribution denoted $\chi$ on $R$ then involves simply generating each coefficient of $x^n, 0 \le n \le 2^{d-1}-1,$ from a bounded discrete Gaussian distribution.  This simple sampling procedure arises due to the modulo $\Phi_{2^d}(x)$, which ensures that the coefficients are all independent after modular reduction.  Reducing modulo an arbitrary monic irreducible polynomial can introduce dependencies between coefficients which ceases to satisfy the assumptions underlying the hardness results of ring-LWE \citep{ringlwe}, leading to more complex sampling procedures.

If $\underline{a}$ is a uniform random draw from $R_q$ this is denoted $\underline{a} \sim R_q$, or correspondingly if $\underline{a}$ is a draw from the multivariate bounded discrete Gaussian draw induced on $R$, $\chi$, this is denoted $\underline{a} \sim \chi$.

\subsubsection{The encryption scheme}
\label{sec:theencryptionscheme}

The message space of this scheme is the polynomial ring $M = R_t$.  Thus any integer message $m$ must be converted to a polynomial representation $\mathring{m}(x)$.  In principle, if $m$ is small enough that $m \in \mathbb{Z}_t$, then the degree zero polynomial $\mathring{m}(x) = m \in R_t$ is sufficient.  However, there are reasons which will become apparent that this is undesirable even when $m$ is small enough (or $t$ is large enough).

A better approach is to take an integer to be encrypted, write it in standard $b$-bit binary representation, $m=\sum_{n=0}^{b-1} a_n 2^n$, and then simply construct $\mathring{m}(x) = \sum_{n=0}^{2^{d-1}-1} a_n x^n \in R_t$ where $a_n = 0 \ \forall\ n \ge b$.  Recovery of the original message after decryption is then simply evaluation of $\mathring{m}(2) = m$, because homomorphic addition and multiplication operations will correspond to operations on the polynomials preserving the end result.  This representation is assumed here and is used automatically in the software contribution of Section \ref{sec:Rpkg}.

The cipher text space is the Cartesian product of two polynomial rings $C = R_q \times R_q$, where $q \gg t$.  As will be seen, the message polynomial is essentially embedded in the $\log_2(t)$ most significant bits of the first polynomial in $C$, with the random noise growing from the least significant bits.  Once the noise grows under repeated operations and reaches the $\log_2(t)$ most significant bits the message is lost.

The parameters of the scheme are: $d$, determining the degree of both the polynomial rings $M$ and $C$; $t$ and $q$, determining the coefficient sets of the polynomial rings $M$ and $C$; and $\sigma$, determining the magnitude of the randomness used for semantic security.

An example of values which ensure good security would be $d=13$ ($\implies 4095$ degree polynomials), $q=2^{128}$, $t=2^{15}$, $\sigma=16$ \citep{Fan12}.  The software contribution of Section \ref{sec:Rpkg} provides functions to help select these parameters automatically based on lower bounds of security and computability they provide.

\noindent\textbf{Key Generation:}
The secret key, $\underline{k}_s$, is simply a uniform random draw from $R_2$ (i.e.\ sample a $2^{d-1}$ binary vector for the polynomial coefficients).

The public key, $\vec{k}_p$, is a vector containing two polynomials: \[ \vec{k}_p = (\underline{k}_{p1}, \underline{k}_{p2}) := ([-(\underline{a} \cdot \underline{k}_s + \underline{e})]_q, \underline{a}) \in R_q \times R_q \] where $\underline{a} \sim R_q$ and $\underline{e} \sim \chi$.  Note $\underline{k}_s$ is hard to extract from $\vec{k}_p$ precisely due to ring LWE hardness results (see Appendix \ref{sec:ringlwe}).

\noindent\textbf{Encryption, $\Enc(\vec{k}_p, m)$:}
An integer message $m$ is first represented as $\underline{\mathring{m}} \in R_t$ as described above.  Encryption then renders a cipher text which is a vector containing two polynomials: \[ \vec{c} = (c_1, c_2) := ([\underline{k}_{p1} \cdot \underline{u} + \underline{e}_1 + \Delta \cdot \underline{\mathring{m}}]_q, [\underline{k}_{p2} \cdot \underline{u} + \underline{e}_2]_q) \in R_q \times R_q \] where $\underline{u}, \underline{e}_1, \underline{e}_2 \sim \chi$ and $\Delta = \left\lfloor \frac{q}{t} \right\rceil$.

\noindent\textbf{Decryption, $\Dec(k_s, \vec{c})$:}
Decryption of a cipher text $c$ is by evaluating:
\[ \underline{\mathring{m}} = \left[ \left\lfloor \frac{t[\underline{c}_1 + \underline{c}_2 \cdot \underline{k}_s]_q}{q} \right\rceil \right]_t \in R_t \]
so that $m = \mathring{m}(2)$.

\noindent\textbf{Addition, $+$:}
Addition in message space is achieved in cipher text space by standard vector and polynomial addition with modulo reduction:
\[ \vec{c}_1+\vec{c}_2 = ( [\underline{c}_{11} + \underline{c}_{21}]_q, [\underline{c}_{12} + \underline{c}_{22}]_q ) \]
It is an easy and enlightening exercise to verify by hand that $\Dec(k_s, \vec{c}_1+\vec{c}_2)$ renders $\underline{\mathring{m}}$.

\noindent\textbf{Multiplication, $\times$:} Multiplication in message space produces a more complex operation in cipher text space which increases the length of the cipher text vector:
\[ \vec{c}_1 \times \vec{c}_2 = \left(
  \left[ \left\lfloor \frac{t(\underline{c}_{11} \cdot \underline{c}_{21})}{q} \right\rceil \right]_q,
  \left[ \left\lfloor \frac{t(\underline{c}_{11} \cdot \underline{c}_{22} + \underline{c}_{12} \cdot \underline{c}_{21})}{q} \right\rceil \right]_q,
  \left[ \left\lfloor \frac{t(\underline{c}_{12} \cdot \underline{c}_{22})}{q} \right\rceil \right]_q
  \right) \]
Although it is still possible to recover $\underline{\mathring{m}}$ from one of these larger cipher texts by modifying the decryption function to be $\left[ \left\lfloor \frac{t}{q}[\underline{c}_1 + \underline{c}_2 \cdot \underline{k}_s + \underline{c}_3 \cdot \underline{k}_s \cdot \underline{k}_s]_q \right\rceil \right]_t$, it is preferable to perform a `relinearisation' procedure which compacts the cipher text to a vector of two polynomials again and reverts to the original decryption procedure.  Thus in practice multiplication is a two step procedure: cipher text multiplication followed by relinearisation.  Description of relinearisation is beyond the scope of this review, but full details are in \citet{Fan12} and it is seamlessly implemented in the software contribution described in Section \ref{sec:Rpkg}.

\subsubsection{A practical note}
\label{sec:APracticalNote}

Above, a binary polynomial representation of integers was proposed as being preferable to a scalar (zero degree polynomial) representation (i.e.~a natural number), even when the message is small enough that $m \in \mathbb{Z}_t$, the reason for which should now be clearer.

Consider the addition operation with the example parameters given above, recall that each coefficient of $\mathring{m}(x)$ must lie in the range $-16,383$ to $16,384$ after computation in order to decrypt correctly, and note that the addition operation results in direct addition of coefficients in the polynomial representations.  Now, bearing these points in mind, if $\mathring{m}(x)=m$ then addition will only render the correct answer so long as the overall final result also remains in the range $-16,383$ to $16,384$.  However, with a binary representation the largest coefficient of any term in $\mathring{m}(x)$ will be $\pm 1$, so that at least $16,384$ additions (possibly more) can be performed and still guaranteed to decrypt correctly, furthermore allowing the final result, $\mathring{m}(2)$, to be much larger than $\pm 16,384$.  Not only is this more additions, but more importantly the binary representation allows a general hard bound for how many additions can be performed while still guaranteeing the correct value is decrypted, without knowledge of the messages.

\subsection{Some limitations}
\label{sec:limitations}

At this juncture it is important to temper any building excitement.  Although \citet{Gentry09} theoretically provided an exemplar for how fully homomorphic schemes could be constructed, the extraordinary theoretical possibilities are constrained by practical limitations.  These crucial limitations mean that it is not simply a matter of taking any algorithm and converting it to run on encrypted data, so that many statistical algorithms are in fact beyond the computational reach of existing homomorphic schemes.

The limitations discussed now are in general common to all current homomorphic schemes to a varying degree, though specific homomorphic encryption algorithms may have their own additional constraints.  In each case, the limitation will be highlighted in the context of the scheme described in Section \ref{sec:FandV}.

\subsubsection{Message space}
\label{sec:Message space}

There are currently no schemes which will directly encrypt arbitrary values in $\mathbb{R}$.  Indeed, the most common message space is simply binary, $M = \{0,1\}$, with this being of particular appeal to theoretical cryptographers because it corresponds to construction of arbitrary Boolean circuits and allows all the results in computational complexity theory to be applied to determine computability.  However, from a practical standpoint this is not presently a very feasible avenue.

However, there are schemes which have an expanded message space, such as $M = \mathbb{Z}/n\mathbb{Z}$, or $M = \{-n, -n+1, \dots, n-1, n\}$ for some integer $n$.  These schemes generally correspond to integer rings or fields (for prime $n$) where ordinary rules of arithmetic can be assumed when results are bounded by $n$.  In many schemes which support expanded message spaces, increasing $n$ will impact the capabilities of the scheme (decreasing security, computation speed, computational depth or all these).

A method which can be used to increase the size of the message space is via the Chinese Remainder Theorem as a means of representing a large integer.

\noindent \textbf{Chinese Remainder Theorem} \citep[p.270]{knuth1997} Let $m_1, \dots, m_k \in \mathbb{Z}^+$ be pairwise coprime positive integers.  Let $M = \prod_{i=1}^k m_i$ and let $a, x_1, \dots, x_k \in \mathbb{Z}$.  Then there is exactly one integer $x$ that satisfies the conditions:
\[ a \le x < a+m \qquad\mbox{and}\qquad x \equiv x_i \mod m_i \ \ \forall\,1 \le i \le k \]

Thus, an integer message $x \in [a, a+m)$ can be uniquely represented by the collection of smaller integers $\{x_i\}_{i=1}^k$, called the residues.  More formally, $\mathbb{Z}/M \cong \mathbb{Z}/m_1 \times \dots \times \mathbb{Z}/m_k$.  So, if each $m_i$ is chosen small enough that the scheme can encrypt it, then much larger message spaces can be achieved by encrypting the collection of residues.  The process is reversible so that the value $x$ can be recovered given $\{x_i\}_{i=1}^k$ \citep[p.274]{knuth1997}.  Such a representation is called a residue number system \citep{RNS} and has the additional advantage that addition and multiplication operations (the only ones which can be performed homomorphically anyway) are embarrassingly parallel: performing the same operation according to the modular arithmetic of each residue will result in a residue representation of the corresponding result of operating on the large integers.

Related and more common in the homomorphic encryption literature, is the reverse usage of the polynomial version of the Chinese Remainder Theorem, which enables combining multiple messages into a single polynomial representation (that is, $\underline{\mathring{m}}$ now holds multiple plain text messages before encryption), so that operations on the single cipher text performs simultaneous operations on all the messages simultaneously in a manner akin to Single Instruction Multiple Data (SIMD) instructions on a CPU \citep{smart14simd}.  This of course reduces rather than increases the possible range of individual messages which can be encrypted.

Even if using the Chinese remainder theorem to represent larger values, the issue remains of how to handle statistical data, which is commonly not binary or integer.  There are at least two approaches: the first is common throughout the literature, whereby any real value is approximated by some rational number, with numerator and denominator encrypted separately and propagated through using the usual rules of arithmetic for fractions.  The second is a logarithmic representation developed by \citet{franz2010secure}, in which division is possible but where addition and subtraction become substantially more complex to implement.

The \texttt{FandV} scheme has an unusual message space, being a polynomial ring.  For the example parameter values given above, this means that when using the binary representation of integer values, the integers can in principle be very large (over $\pm 10^{1237}$).  As such, the limitation in message space size may seem less acute than in other homomorphic schemes (especially binary ones), but the practical issue raised in \S\ref{sec:APracticalNote} means that it may still be advantageous to use a residue number system representation if there will be a lot of addition. 

In the follow on to this review \citep*{Part2}, two other approaches are proposed: one where data is effectively quantile binned in a binary indicator fashion, which is shown to effectively enable simple comparison operations; and another discretisation of real values which is appropriate for linear modelling.

\subsubsection{Cipher text size}

Once the value to be encrypted has been appropriately represented such that only elements of $M$ need to be encrypted, there is the additional issue of a substantial inflation in the size of the message after encryption, often by several orders of magnitude.

As a concrete example, the usual representation of an integer in a computer requires 4 bytes of memory.  If such a message is encrypted under the scheme presented in Section \ref{sec:FandV}, then using the example parameters will result in cipher texts occupying $65,536$ bytes (4096 coefficients, each a 128-bit integer).  Consequently, a 1MB data set will occupy nearly 16.4GB encrypted.

One mitigating proposal \citep{naehrig2011can} is to initially encrypt values using a non-homomorphic, size efficient encryption algorithm such as AES, and to encrypt the AES decryption key with a homomorphic scheme.  The decryption circuit for AES can then be executed homomorphically, rendering a homomorphic encryption of the original message.  This would mean that communication and long term storage of encrypted values could be space efficient, with expanded homomorphic cipher texts generated by effectively `recrypting' from this compact format when computation is required.  AES is an industry standard, but required 36 hours to execute homomorphically \citep{gentry2012homomorphic} (for 56 AES blocks, corresponding to 896 bytes of data), although a more recent lightweight cipher named SIMON can be recrypted homomorphically in around 12 minutes \citep{lepoint2014comparison}.  However, these approaches operated on binary messages, so the resulting recryption is to a binary scheme with the attendant issues already discussed.

\subsubsection{Computational cost}

Elements of cipher text space are not only larger in memory (with an associated additional computational cost to process), but will typically also be more complex spaces.  For example, in Section \ref{sec:FandV} the cipher text space is the ring of polynomials modulo a cyclotomic polynomial, with coefficients from a large integer ring (e.g.~128-bit integers).  Consequently, arithmetic operations are substantially more costly than standard arithmetic: there is large polynomial arithmetic involving coefficients which are too large to fit in standard 32-bit or 64-bit integers, with the additional overhead of modulo operations on both the coefficients and polynomial.

Most current schemes can achieve reasonable speeds for additions, but are very constrained in speed of multiplications.  The optimised scheme implemented in the R package \texttt{HomomorphicEncryption} \citep{HEpkg} achieves thousands of additions per second, and about 50 multiplications per second.  This is mitigated as far as possible by transparently implementing full CPU parallelism.

If all the operations involved can be performed in a single instruction multiple data (SIMD) fashion then the polynomial Chinese remainder theorem alluded to above can be used when representing the messages as a polynomial prior to encryption.  In this way a single cipher text operation actually operates in a SIMD manner on many messages for the same computational cost \citep{smart14simd}.  Naturally, there is a limit to how many messages can be packed into a single cipher text in this way.

\subsubsection{Division and comparison operators}
\label{sec:div}

At present there are no homomorphic schemes capable of natively supporting division operations, only addition and multiplication.  An additional serious constraint is the inability to have any conditional code flow: comparison operators such as tests of equality and inequality cannot be performed on the encrypted data.  Consequently, many algorithms appear out of reach without substantial redevelopment.

\subsubsection{Depth of operations}

The final limitation relates to the number of operations which can be applied.  As explained in the discussion on semantic security, there is randomness injected into the cipher text in these encryption schemes.  When operations are performed, the noise tends to accumulate (exactly how being scheme dependent): for example, in many schemes multiplication operations result in direct multiplication of the noise components leading in the na\"{i}ve case to potentially exponential increases in the magnitude of the noise over many operations.  Once the noise exceeds a certain threshold then decryption will render the incorrect message.

It is important to be clear that it is not usually the total number of multiplications which is limited, but rather the \emph{depth} (i.e.\ the maximum degree of the evaluated polynomial).  For example, $x_1\times x_2\times x_3$ has multiplicative depth 2, whereas $x_1 \times x_2 + x_3 \times x_4 + \dots + x_{n-1} \times x_n$ has multiplicative depth 1 $\forall\,n$.  Exactly what depth a scheme can achieve will depend on the scheme itself and usually on the parameters chosen, which commonly involves a tradeoff of speed, security or memory requirements against depth of operations.

In principle, one of the breakthrough aspects of \citepos{Gentry09} work was the ability to \emph{bootstrap} (entirely unrelated to the statistics term) a cipher text: an operation which resets the noise to that of a freshly encrypted message.  However, most bootstrapping routines are very complex to implement, extremely slow to execute, or both.  As a result, it is almost universal in the applied cryptography literature to set the parameters of the scheme under consideration to be such that the necessary depth of operations can be performed without a bootstrapping step being required.  The software contribution of Section \ref{sec:Rpkg} provide functions to help automatically select the parameters based on lower bounds in the literature for the depth of multiplications required.

\subsubsection{Motivation}

To date the small number of applied cryptography papers have largely taken existing statistical techniques which can be made to directly fit within these constraints and demonstrated any minor refactoring of the algorithms that is necessary, but leave them fundamentally unaltered (some examples are reviewed in Section \ref{sec:CurrentMethods}).  However, statisticians and machine learners are well placed to develop principled approximations to current statistical and machine learning techniques, or entirely new techniques, where the constraints of homomorphic encryption are considered at all stages of model and algorithm development, and where uncertainties and errors introduced can be studied.  Some initial contributions in this direction are presented in \citet*{Part2}.

\subsection{Usage scenarios}

The most obvious usage scenario is to outsource long-term storage and computation of sensitive data to a third party cloud provider.  Here the `client' (the owner of the data) encrypts everything prior to uploading to the `server' (at the cloud provider's data centre).  Due to some of the limitations discussed above, this scenario is perhaps currently only suitable in a restricted set of situations where the added computational costs and inflated data size are not prohibitive.  With homomorphic schemes improving all the time the boundary where this is a practical usage scenario will shift over time.

However, with the explosion of extremely compute, memory and battery constrained devices such as smart watches and glasses it may be that scenarios where additional server side memory and compute costs are a worthwhile trade-off are substantially broader.  This is especially true given the biomedical focus of many of these recent devices which collect a lot of sensitive health data: collection of this on constrained client devices and handoff to a cryptographically secure server storage area which is capable of encrypted statistical analysis is an attractive proposition for both users and manufacturers.

An additional scenario is one in which it is desirable to be able to perform statistical analyses without the data being visible to anyone at all.  To be concrete, consider a research institute requiring patient data for analysis: the research institute could widely distribute their public key to enable patients to securely donate their sensitive personal data.  This data would be encrypted and sent directly to the cloud provider who would have a contractual obligation to only allow the research institute access to the results of pre-approved functions run on that data, not to the raw encrypted data itself.  Peer review would be important for pre-approving certain functions to be homomorphically executed to ensure that the original data is not indirectly leaked.  An interesting effect here may be increased statistical power (despite homomorphic approximations) due to the greater sample sizes which could result from increased participation because of the privacy guarantees.

There is at least one further usage scenario: that is, where there is confidential data on which a confidential algorithm must be run.  In this situation, a client may encrypt their data to give to the developer of the algorithm and receive the results of the algorithm without either party compromising data or algorithm.  In this situation, the constraints of homomorphic encryption are merely an opportunity cost because there may be no other way to achieve the same goal.

\section{Current Methods}
\label{sec:CurrentMethods}

There are two aspects which, from the perspective of a statistician, are important to review: prior work on encrypted statistics algorithms and existing software implementations for making use of homomorphic encryption schemes.

In this section, both aspects are surveyed before the software tools documented in this paper are covered in Section \ref{sec:Rpkg}.

\subsection{Encrypted statistics}

In the recent years, some work has emerged on statistical methods for homomorphically encrypted data.

\citet{GLN12}
proposed algorithms for binary classification, namely secure versions of the Linear Means and Fisher's Linear Discriminant classifiers.
The algorithms are rewritten in such a way that divisions are avoided but the original score function (needed for classification) is computable up to a constant. Because some operations have no counterpart in the encryption framework (like division and comparison), some of the computation is done offline by the client after decrypting results returned by the cloud. For instance, in binary classification $y \in \{-1,1\}$ with Linear Means, the class label is computed in this way as the sign of a score function.
To represent real numbers as integers, the authors propose a rescaling approach which approximates real numbers with rational numbers  (integer numerator and denominator) and then clears denominators by multiplying all numbers by an appropriate factor and rounding the result to the nearest integer. Approximation accuracy can be controlled in this way.

\citet{WH12}
extended previous work on encrypted statistics \citep{Lauter11}, namely the computation of mean and covariance in a multivariate scenario, using the same technique of returning separate encrypted numerators and denominators.
Additionally, they also mention the possibility of implementing (and indeed implement) low-dimension linear regression ($d \leq 5$) by using Cramer's rule to invert the matrix $X^{T}X$ which is required to obtain the ordinary least squares estimates of the regression parameters $\hat{\beta} = (X^{T}X)^{-1}X^{T}Y$.
Because Cramer's rule also involves a division by the determinant of $X^{T}X$, the computation can not be completely performed homomorphically and must be finished offline by the client who assembles the division factors post-decryption.
Apart from the computational issues caused by division, there are additional problems here, the most important being the complexity of Cramer's rule: for a problem with dimension $d$, the computation of the determinant has multiplicative depth $d-1$ and requires $\mathcal{O}(d!)$ multiplications. Allied to this comes the computation of the adjoint matrix, having similarly substantial computational complexity.
The restriction is two-fold: firstly, in the multiplicative depth of operations; and secondly, in the computational costs of these operations. Whereas the second restriction implies possible intractability of high-dimensional linear regression, the first restriction affects correctness of decryption and so should be regarded as more serious.

\citet{LLN14}
observed that it is possible to analyse genomic data in a privacy-preserving framework and provide some examples of algorithms in statistical genetics which are implementable under the restrictions of homomorphic encryption, including the Cochran--Armitage trend test, the expectation--maximisation algorithm and measures of goodness-of-fit and linkage disequilibrium.
The main issue in implementing these methods under the homomorphic encryption framework is that divisions are not possible. The solution proposed is to write the statistics in terms of the two factors involved in a division (dividend/numerator and divisor/denominator), compute these homomorphically and send them back to the client, who decrypts each factor and performs the division offline.
For complex problems where divisions can not be grouped (by combining dividends and divisors), there will be a higher number of cipher texts being passed to the client, which increases communication costs and, more importantly, may compromise privacy since more information is contained in less processed cipher texts.

Another class of privacy-preserving statistical methods has been proposed for predictive purposes: an algorithm is trained offline (say, a regression model) and the corresponding predictive model (the parameters in the regression model, $\beta$) encrypted.
For prediction tasks, covariates are encrypted and sent to the server, where computations take place (e.g., the computation of the regression model predictor, $X_{i\smallbullet}\beta$) and are then returned to the client for decryption (and potentially further transformation, as would be the case for generalised linear models).
Examples of these include logistic regression \citep{yashe} and hidden Markov models \citep{PRSR11}

Crucially, in all these current methods, existing algorithms are simply refactored to run homomorphically rather than developing novel approaches to approximate otherwise currently intractable statistical techniques. 

\subsection{Implementations}

As will be clear from Section \ref{sec:FandV}, many homomorphic schemes can be non-trivial to implement.  Some public implementations are releases of software which was written for a specific paper, whilst there are a small number of libraries or packages enabling reuse.  Most libraries or packages commonly interfaces in low-level languages such as C/C++.  A very compact single C file library implementing \citet{Gentry10} is `libfhe' \citep{libfhe}.  This implementation is based on a binary scheme, but has routines to allow encryption of integers by base-2 decomposing, encrypting each binary digit separately and then implementing binary adder arithmetic (so that even addition will involve cipher text multiplications).  There is no bootstrapping implementation and at time of writing there have been no apparent updates since 2010.

`Scarab' \citep{scarab} is another low-level C library, implementing instead another integer cipher text space scheme by \citet{smart2010fully}.  This implementation allows only encryption of a binary message, although as well as providing addition (XOR) and multiplication (AND), there are full and half adders provided offering carry in and carry out or just carry out, respectively.  A bootstrapping routine is also provided.  There have not been additional updates in some time.

Another low level implementation, `HELib' \citep{helib}, provides a C++ library implementing \citet{brakerski2012leveled}, one of the early second generation of schemes (see Appendix \ref{sec:modernHE}).  It incorporates some very useful optimisations, including the work of \citet{smart14simd}, which enables single-instruction multiple-data (SIMD) parallelism by packing multiple values in a single cipher text.  This is under active development at the time of writing and appears the most comprehensive implementation of a modern scheme currently available.  Details of the algorithms used are available in preprint \citep{helib2}.

Finally, there was a recent comparison of two schemes, \citet{Fan12} and \citet{yashe}, in \citet{lepoint2014comparison} which provided the C++ software used \citep{homomorphic-simon}.  Although not in the explicit form of a library it could be possible to transform this into a C++ library for the two schemes.

\section{\texttt{HomomorphicEncryption} R package}
\label{sec:Rpkg}

For statistics researchers to be able to use homomorphic encryption techniques, an easy to use yet high performance library in a high level language which is popular in the community is necessary.  
An R language \citep{R} package providing such an implementation is a contribution of our work.

The \texttt{HomomorphicEncryption} R package \citep{HEpkg} provides an easy to use interface to begin developing and testing statistical methods in a homomorphic environment.  The package has been developed to be extensible, so that as new schemes are researched by cryptographers they can be made available for use by statistics researchers with minimal additional effort.  The package has a small number of  generic functions for which different cryptographic backends can be used.  The underlying implementation is mostly in high performance C and C++ \citep{Rcpp}, with many of the operations setup to utilise multi-core parallelism via multithreading \citep{RcppParallel} without requiring any end-user intervention.

The first generic cryptographic function is \texttt{pars}.  The first argument to this function designates which cryptographic backend to use and allows the user to override any of the default parameters of that scheme (for example, $d, q, t$ and $\sigma$ of Section \ref{sec:FandV}).  Related to this, there is the alternative method of specifying parameters via the function \texttt{parsHelp}.  This allows users to instead specify a desired minimal security level in bits and a minimal depth of multiplications required, and then computes values for $d, q, t$ and $\sigma$ which will satisfy these requirements with high probability, by automatically optimising established bounds from the literature \citep{lepoint2014comparison, lindner2011better}

The second generic cryptographic function is \texttt{keygen}, whose sole argument is a parameter object as returned by \texttt{pars} or \texttt{parsHelp}.  \texttt{keygen} then generates a list containing public (\texttt{\$pk}) and private (\texttt{\$sk}) keys, along with any scheme dependent keys (such as relinearisation keys in the case of \citet{Fan12}), which correspond to the homomorphic scheme designated by the parameter object.  At this point, the parameter object is absorbed into the keys so that it doesn't need to be used for any other functions.

The third generic cryptographic function is \texttt{enc}.  This requires simply the public key (as returned in the \texttt{\$pk} list element from \texttt{keygen}) and the integer message to be encrypted.  It then returns a cipher text encrypted under the scheme to which the public key corresponds.  Crucially, the ease of use begins to become very apparent here, with \texttt{enc} overloaded to enable encryption of not just individual integers, but also vectors and matrices of integers defined in R.  The structure of the vectors and matrices are preserved and the encryption process is fully multithreaded across all available CPU cores automatically.

The final generic cryptographic function is \texttt{dec}.  Similarly, this requires simply the private key, as returned in the \texttt{\$sk} list element from \texttt{keygen}, and the (scalar/vector/matrix) cipher text to be decrypted.  It then returns the original message.  Note that the structure of vector or matrix cipher texts is correctly preserved throughout.

The real simplicity becomes evident when manipulating the cipher texts.  All the standard arithmetic functions (\texttt{+}, \texttt{-}, \texttt{*}) work as expected, implementing for example the cyclotomic polynomial ring algebra of the \texttt{FandV} scheme transparently.  Moreover, vectors can be formed in the usual R manner using \texttt{c} (or extracted from the diagonal of matrix cipher texts with \texttt{diag}), element wise arithmetic can be performed on those vectors (with automatic multithreaded parallelism) and there is support for all the standard vector functions, such as \texttt{length}, \texttt{sum}, \texttt{prod} and \texttt{\%*\%} for inner products, just as one would conventionally use with unencrypted vectors in R.  Indeed, such functionality extends to matrices, with formation of diagonal matrices via \texttt{diag} from cipher text vectors, element wise arithmetic and full matrix multiplication using the usual \texttt{\%*\%} R operator (again, automatically fully parallelised).  Matrices also support the usual matrix functions (\texttt{dim}, \texttt{length}, \texttt{t}, etc).  The package automatically dispatches these operations to the correct backend cryptographic routines to perform the corresponding cipher text space operations transparently, returning cipher text result objects which can be used in further operations or decrypted.

The following is the simplest possible instructive example.  Examining the contents of \texttt{k}, \texttt{c1}, etc will show the encryption detail:
\begin{Shaded}
\begin{Highlighting}[]
\KeywordTok{library}\NormalTok{(HomomorphicEncryption)}
\NormalTok{p <-}\StringTok{ }\KeywordTok{pars}\NormalTok{(}\StringTok{"FandV"}\NormalTok{)}
\NormalTok{k <-}\StringTok{ }\KeywordTok{keygen}\NormalTok{(p)}
\NormalTok{c1 <-}\StringTok{ }\KeywordTok{enc}\NormalTok{(k$pk, }\KeywordTok{c}\NormalTok{(}\DecValTok{42}\NormalTok{,} \DecValTok{34}\NormalTok{))}
\NormalTok{c2 <-}\StringTok{ }\KeywordTok{enc}\NormalTok{(k$pk, }\KeywordTok{c}\NormalTok{(}\DecValTok{7}\NormalTok{,} \DecValTok{5}\NormalTok{))}
\NormalTok{cres1 <-}\StringTok{ }\NormalTok{c1 + c2}
\NormalTok{cres2 <-}\StringTok{ }\NormalTok{c1 * c2}
\NormalTok{cres1[1]}
\KeywordTok{dec}\NormalTok{(k$sk, cres1)}
\KeywordTok{dec}\NormalTok{(k$sk, cres2)}
\end{Highlighting}
\end{Shaded}

Note that indexing into vectors and matrices as provided by R via the usual \texttt{[]} notation is fully supported, including assignment.

We hope this provides a distinctly easy-to-use software implementation in arguably the most popular high level language in use among data scientists today, including automatic help for encryption scheme parameter selection to aid non-cryptographers.  Moreover, given the computational burden of homomorphic schemes, the transparent multithreaded parallelism automatically across all CPU cores in all available scenarios (encryption, decryption and arithmetic with vectors/matrices) enables focus to be on the subject matter questions.

At present, the scheme of \citet{Fan12} (described in Section \ref{sec:FandV}) has been implemented, making use of FLINT \citep{Hart2010} for certain polynomial operations and GMP \citep{gmp} for high performance arbitrary precision arithmetic.  Backends for further homomorphic encryption schemes may be added in the future.

Table \ref{tab:timezzz} provides indicative timings for common operations using the default parameters of the package (which match the default parameters suggested in \citealp{Fan12}).

\begin{table}[!h]
\captionsetup{width=0.76\textwidth}
\caption{Timings (in seconds; average of 100 repetitions) for operations on cipher texts using the \texttt{HomomorphicEncryption} package.  All timings performed on an Amazon EC2 c4.8xlarge instance for reproducibility. $S$ represents a scalar, $V$ a vector of size 100 and $M$ a matrix of size $10 \times 10$.\hfill{ }}\vspace{-20pt}
\label{tab:timezzz}
\begin{small}
\begin{center}
\begin{tabular}{cccccccccccc}
\hline\hline
&\multicolumn{2}{c}{\textbf{scalar operations}} && \multicolumn{2}{c}{\textbf{vector operations}} && \multicolumn{2}{c}{\textbf{matrix operations}} \\
\cline{2-3}\cline{5-6}\cline{8-9}
&$S\texttt{+}S$       & 0.003 &&
 $V\texttt{+}V$       & 0.58    &&
 $M\texttt{+}M$       & 0.87    &\\
&$S$\texttt{*}$S$     & 0.084 &&
 $V$\texttt{*}$V$     & 1.59    &&
 $M$\texttt{*}$M$     & 8.49    &\\
&                                         &             &&
 $V$\texttt{\%*\%}$V$ & 1.59    &&
 $M$\texttt{\%*\%}$M$ & 10.21 &\\
\hline
\end{tabular}
\end{center}
\end{small}
\end{table}

\section{Conclusions}

This technical report has provided a review of homomorphic encryption with a focus on issues which are pertinent to statisticians and machine learners.  It also introduces the \texttt{HomomorphicEncryption} R package and demonstrates the ease of getting started experimenting with homomorphic encryption.

The practical limitations of homomorphic encryption schemes means that existing techniques cannot always be directly translated into a corresponding secure algorithm.  This presents an opportunity for the statistics and machine learning community to engage with research in privacy preserving methods by developing new methods which are tailored to homomorphic computation and which work within the constraints described in Section \ref{sec:limitations}, with the sister paper to this review \citep*{Part2} being an initial contribution in this direction.

\section*{Acknowledgements}

The authors would like to thank the EPSRC and LSI-DTC for support.  Louis Aslett and Chris Holmes were supported by the i-like project (EPSRC grant reference number EP/K014463/1).  Pedro Esperan\c{c}a was supported by the Life Sciences Interface Doctoral Training Centre doctoral studentship (EPSRC grant reference number EP/F500394/1).

\renewcommand\bibname{References}
\bibliography{FHE_paper}
\bibliographystyle{Louis_agsm}

\appendix
\section{Modern homomorphic schemes}
\label{sec:modernHE}

The groundbreaking work by \citet{Gentry09} set the stage for the modern era of homomorphic schemes where both addition and multiplication to a (theoretically) arbitrary depth are possible.  In a nut shell, Gentry constructed a scheme based on ideal lattices over a polynomial ring which could perform sufficient homomorphic operations to evaluate a so-called `squashed' version of its own decryption algorithm: thus, given an encrypted version of a hint about the secret key, evaluating the decryption homomorphically results in a `fresh' cipher text where the noise level is reset.

This quickly spawned many other schemes which invoked these techniques.  Two conceptually much simpler schemes using the technique and based on large integer cipher texts were developed in \citet{van2010fully} and \citet{smart2010fully}.  \citet{stehle2010faster} directly improved on \citet{Gentry09} making evaluation of operations less complex.  \citet{brakerski2011fully} used the Gentry approach removing some untested security assumptions which had been made.  These works were in a sense the `first generation' of modern schemes.

\citet{brakerski2011efficient} triggered a second generation of schemes based on the ``learning with errors'' (LWE) problem \citep{regev2009lattices} which did not rely on the poorly understood hardness assumptions of ideal lattices or `squashing' of the decryption circuit to achieve full homomorphism.  Moreover, it ensured that the size of the public key was independent of the depth of operations to be performed: implementations of Gentry's original scheme required upto 2.3 gigabyte public keys \citep{gentry2011implementing}!  This second generation of schemes includes \citet{brakerski2012leveled} which introduced `leveled' schemes, where noise grows linearly; \citet{brakerski2012fully} which introduced scale-invariance reducing the number of keys that must be stored; \citet{Fan12} which provided a practical scheme, porting scale invariance to the \citet{brakerski2012leveled} scheme and setting it in a ring-LWE context \citep{ringlwe}; \citet{gentry2013homomorphic} which introduced a highly novel LWE approach where cipher texts are matrices and operations follow standard matrix arithmetic; and \citet{brakerski2014lattice} where they focus on matching security levels of non-homomorphic schemes, among others.

\section{Ring Learning With Errors (LWE)}
\label{sec:ringlwe}

The ring LWE hardness result underlies the homomorphic encryption scheme reviewed in Section \ref{sec:FandV}.  It is a ring based extension of the original LWE result due to \citet{regev2009lattices}.  For the interested reader this appendix provides a short simplified explanation of the problem the security of the scheme relies upon. The notation here follows that of Section \ref{sec:FandV}.

The original LWE problem requires reconstruction of a secret vector $\vec{s} = (s_1, \dots, s_n) \in \mathbb{Z}_q^n$, for some $q \in \mathbb{Z}$, when only in possession of a collection of approximate random linear equations.  First, imagine forming the results of many linear equations, $z_j = \langle \vec{a}_j, \vec{s} \rangle$, by choosing uniformly random vectors $\vec{a}_j \sim \mathbb{Z}_q^n$.  Then, given $n$ realisations of $\{ \vec{z}_j, \vec{a}_j \}$ it is a simple matter of solving a system of linear equations to recover $\vec{s}$.

However, consider the approximate version of this problem: given a uniformly random vector $\vec{a}_j \sim \mathbb{Z}_q^n$, form instead the perturbed inner products $\vec{y}_j = \langle \vec{a}_j, \vec{s} \rangle + e_j$ where $e_j$ is a scalar discrete random Gaussian draw.  Then, given many realisations of $\{ \vec{z}_j, \vec{a}_j \}$ the objective is to solve $\langle \vec{a}_j, \vec{x} \rangle \approx \vec{y}_j$ for $\vec{x}$.  For appropriate choices of the error this can be shown to be an exceptionally hard problem: certainly as hard as traditional worst-case lattice problems which have been well studied.

Ring LWE \citep{ringlwe} ports the same results to the more complex polynomial ring setting, but the formulation is essentially unchanged in that it is now simply solution of a system of perturbed linear equations in an algebraic ring.

Notice that the public key in Section \ref{sec:FandV} is precisely the ring LWE problem: the public key contains a masked version of the secret key, with the security of doing this based on the difficulty of recovering it due to the ring LWE problem hardness.

\end{document}